\documentclass{article}
\usepackage{spconf,amsmath,graphicx}
\usepackage{hyperref}
\usepackage{algorithm}
\usepackage{algorithmic}
\usepackage{booktabs}
\usepackage{multirow}
\usepackage{textcomp}
\usepackage{stfloats}
\usepackage{url}
\usepackage{verbatim}
\usepackage{cite}

\usepackage{makecell}
\usepackage{subfig}
\usepackage{float}                  
\usepackage{overpic}
\title{Leveraging Motion Estimation for Efficient Bayer-Domain Video Convolutional Networks}
\name{Haichao Wang\textsuperscript{1}, Jiangtao Wen\textsuperscript{2}\sthanks{Jiangtao Wen is the corresponding author.}, Yuxing Han\textsuperscript{1}}
\address{\textsuperscript{1}Shenzhen International Graduate School, Tsinghua University, China \\ \textsuperscript{2}New York University, USA}
\begin{document}
%
\maketitle
\begin{abstract}
Existing computer vision pipelines suffer from two efficiency bottlenecks: image signal processors (ISPs) first convert Bayer pixel data to RGB pixel by pixel, and video models repeatedly process temporally redundant frames, leading to high power consumption and latency. In this paper, we propose a novel framework that eliminates the ISP and leverages motion estimation to accelerate video vision tasks directly in the Bayer domain. We introduce Motion Estimation-based Video Convolution (MEVC), which integrates sliding-window motion estimation into each convolutional layer, enabling prediction and residual-based refinement that reduces redundant computations across frames. This design bridges the structural gap between block-based motion estimation and spatial convolution, enabling accurate, low-cost processing. Our end-to-end pipeline supports raw Bayer input and achieves over 70\% reduction in FLOPs with minimal accuracy degradation across video semantic segmentation, depth estimation, and object detection benchmarks, using both synthetic Bayer-converted and real Bayer video datasets.
\end{abstract}
\section{Introduction}
\label{sec:intro}
Video computer vision is a key area of artificial intelligence, supporting applications such as autonomous driving and security systems. In these resource-constrained or real-time scenarios, efficiency remains a challenge even for lightweight CNN-based video vision architectures~\cite{he2016deep}. Specifically, video vision systems are hampered by bottlenecks at both the front-end and back-end. At the front-end, a computationally intensive Image Signal Processor (ISP) converts raw Bayer data into RGB format pixel by pixel before feeding it to vision models. At the back-end, conventional frameworks process high-resolution video frame by frame, failing to exploit inherent temporal redundancy. This approach leads to unnecessarily high computational complexity and latency.

\begin{figure}
    \centering
    \includegraphics[width=1\linewidth]{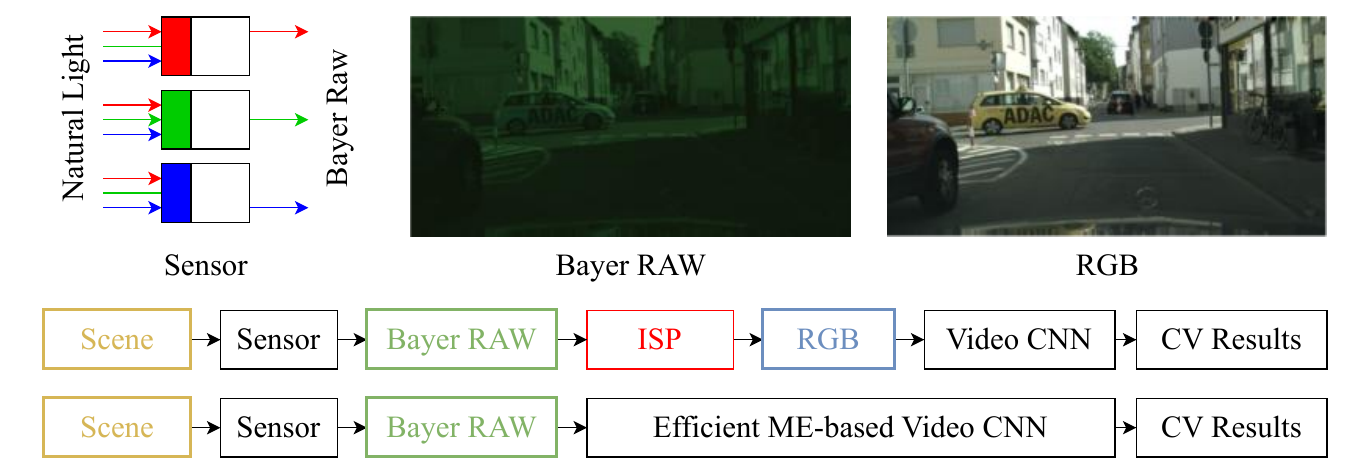}
    \caption{Top: Traditional video computer vision pipeline. Bottom: Our proposed efficient Bayer-domain video computer vision system.}
    \label{fig:intro}
\end{figure}

To improve efficiency, researchers have studied video convolutional network acceleration using two approaches. The first approach processes key frames and non-key frames with different models, using larger models for key frames and smaller models for non-key frames~\cite{jain_accel_2019, hu_efficient_2023}. A compensation module then enhances current-frame outputs using key-frame information. For videos with limited scene changes, such schemes perform reasonably well; however, redundant calculations still exist. The second approach aims to reduce temporal redundancy more directly. For instance, BlockCopy~\cite{verelst_blockcopy_2021} divides frames into blocks and uses a reinforcement learning model to decide whether a current-frame block should be copied from the co-located block in the reference frame or processed by a standard model. By ignoring differences between current and reference blocks, vision performance degrades rapidly as the acceleration ratio increases.

This paper proposes an efficient video computer vision framework to accelerate VCN-based vision with minimal performance degradation. As shown in Figure~\ref{fig:intro}, instead of converting Bayer videos into RGB videos for subsequent processing, we introduce a Bayer-format pipeline that directly uses Bayer data and removes the ISP from the pipeline. Bayer training and testing data are obtained using Invertible-ISP~\cite{xing2021invertible}, an invertible convolutional network with a differentiable JPEG module. We then propose an efficient video convolution called Motion Estimation-based Video Convolution (MEVC), where repeated and unnecessary convolution calculations are reduced by introducing motion estimation (ME) into each convolutional layer. For each block in a current frame, motion estimation finds the most similar block in the reference frame and records the corresponding motion vector and residual. The output of the current frame is predicted from the reference frame with residual compensation using the motion vector (MV) and residual. Since residual maps are sparse, the computation for compensation is greatly reduced. The key challenge lies in the gap between motion estimation and convolution, which sliding-window motion estimation and compensation are designed to address. Our main contributions are: 1) an efficient MEVC operation that is generic to different computer vision tasks and VCNs; 2) an end-to-end Bayer-format pipeline for video vision without ISP processing; 3) sliding-window motion estimation and compensation for MEVC; and 4) extensive experiments on multiple tasks and datasets, including public datasets and a self-collected Bayer-format dataset, demonstrating significant acceleration with minimal performance loss.

\section{Related Works}
Video convolutional networks (VCNs) are widely used in video understanding tasks such as segmentation, detection, and depth estimation. However, their computational cost is high due to repeated processing of temporally similar frames. Prior works on VCN acceleration mainly fall into two categories.

The first category introduces temporal redundancy-aware scheduling~\cite{hu2020temporally}. 
For example, Accel~\cite{jain_accel_2019} and AR-Seg~\cite{hu_efficient_2023} process key frames with a full model and interpolate non-key frames using lightweight models or low-resolution inputs, with feature-level fusion for compensation. While effective in low-motion scenarios, these methods often fail to adapt to varying motion levels and suffer from over-computation on low-dynamic frames or degraded performance on high-motion ones.
The second category attempts to exploit frame-level redundancy more directly. BlockCopy~\cite{verelst_blockcopy_2021} and TapLab~\cite{feng_taplab_2022} employ block-wise reuse of feature maps based on similarity, using learned or heuristic policies to determine when to copy or compute. However, these approaches typically omit residual compensation and are sensitive to alignment errors, leading to noticeable accuracy degradation under large or misaligned motion.

In contrast to both paradigms, our method introduces motion estimation directly into the convolutional layers, enabling fine-grained reuse of computation across frames with explicit residual correction. Unlike methods that reuse entire blocks or interpolate features, our approach decomposes convolution into motion-predicted outputs and sparse residual updates, maintaining accuracy while achieving higher compression.

Recent raw-domain vision studies bypass parts of the conventional ISP by learning directly from raw measurements or by synthesizing Bayer-pattern inputs for downstream tasks~\cite{datta2023enabling, zhou2021raw, rawseg2023, rawadapter2024}. These works mainly study raw-input representation and adaptation. Our work is complementary: we use the Bayer domain as the system input and focus on reducing convolutional computation through motion-compensated reuse.

\section{Methods}
\subsection{End-to-End Bayer-Domain Pipeline}
Conventional computer vision pipelines rely on ISPs to convert raw Bayer sensor data into RGB images. This preprocessing is costly and non-differentiable. To bypass this bottleneck, we directly use Bayer data as input to the neural network and train all models in the Bayer domain. As illustrated in Figure~\ref{fig:pipeline}, video convolution is accelerated using sliding-window motion estimation to identify temporal correlations between blocks in reference and current frames, as well as the residual map, followed by a compensation framework to correct the predicted output based on the residual map.
This design eliminates the computational overhead of ISP while preserving compatibility with standard video vision datasets and tasks.

\subsection{Motion Estimation-based Video Convolution}
\begin{figure}
    \centering
    \includegraphics[width=1\linewidth]{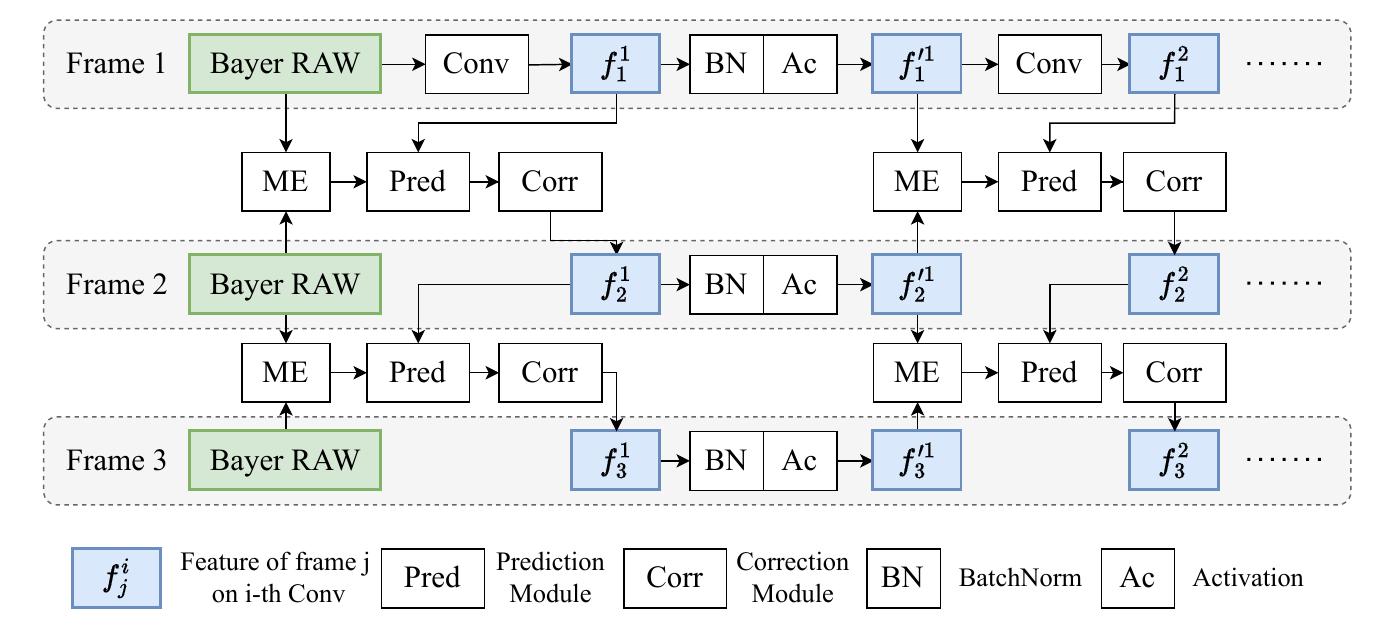}
    \caption{An overall diagram of our framework.}
    \label{fig:pipeline}
\end{figure}

To reduce redundant computation across temporally adjacent frames, we introduce Motion Estimation-based Video Convolution (MEVC), a convolutional module that incorporates motion estimation directly into feature extraction.
Each video sequence is divided into groups of pictures (GOPs). The first frame in each GOP is treated as a key frame and is processed using standard convolutions. For subsequent frames (non-key frames), we apply MEVC modules to reuse computation from previous frames. 

In traditional video coding,  
ME is performed for each block in a current frame to find the best matching block in the reference frame within a search range, where the position difference is recorded as Motion Vector (MV) and the block difference is recorded as Residual Map $\textit{Res}$. During decoding, the current frame $F_t$ is reconstructed by the motion vectors and the residual map from the reference frame $F_{t-1}$.

This process can be expressed as
\begin{gather}
    \textit{MV}, \textit{Res} = \text{ME}\left(F_{t-1}, F_{t}\right) \\
    F_{t}=\textit{predict}\left[F_{t-1},\textit{MV}\right] + \textit{Res}
    \label{eq:videocoding}
\end{gather}
where $F_{t-1}$ denotes the reference frame and $F_{t}$ denotes the current frame.


Convolution is a linear operator, therefore equation~\ref{eq:videocoding} can be expressed as 
\begin{equation}
\textit{Conv}\left(F_{t}\right)=\textit{Conv}\left[\textit{predict}\left(F_{t-1},\textit{MV}\right)\right]+\textit{Conv}\left(\textit{Res}\right)
\label{eq:conv}
\end{equation}
where $\textit{Conv}$ denotes a convolutional operation. Because of the correspondence between the input blocks and the output pixels, equation~\ref{eq:conv} can be further transformed to
\begin{equation}
\textit{Conv}\left(F_{t}\right)=\textit{predict}\left[\textit{Conv}\left(F_{t-1}\right),\textit{MV}\right]+\textit{Conv}\left(\textit{Res}\right) \nonumber
\end{equation}
where the $\textit{Conv}(F_{t-1})$ is already known during the convolution over the reference frame, and residual map $\textit{Res}$ is usually sparse. The above analysis can be generalized to features at different convolutional layers of a convolution network.

FLOPs is a device-independent metric for computation complexity. Assume the input data with shape \( (C_\textit{in}, H_\textit{in}, W_\textit{in}) \) is processed through a 2D convolution layer with \( C_\textit{out} \) filters, a kernel size of \( k \), and a stride of \( s \), resulting in an output shape of \( (C_\textit{out}, H_{\textit{out}}, W_{\textit{out}}) \). For ME, suppose the matching success ratio is called match ratio $\alpha$ and the sparsity of residual map is recorded as $\beta$. 
The acceleration ratio is: $\alpha-\alpha \beta-{(2R+1)^2}/(s^2C_{\textit{out}})$. Here, $\alpha$ and $\beta$ are measured from motion estimation, while $s$ and $C_{\textit{out}}$ are fixed by the convolution layer. The main user-controlled parameter is the search range $R$: a larger $R$ covers larger motion but increases ME cost quadratically through the $(2R+1)^2$ search window.

At a higher level, as shown in Figure~\ref{fig:pipeline}, the proposed framework consists of a standard convolution, ME, prediction, and correction module. Key frames are first processed by the convolution layer, whose output serves as a reference for the prediction module. The prediction module warps the current frame using motion vectors estimated by ME, and the correction module refines it with sparse convolution on residual maps. During deployment, standard convolution is replaced with MEVC to obtain the final models. MEVC only replaces convolution layers; batch normalization and activation layers remain unchanged.

\subsection{Sliding-window Motion Estimation}
In our proposed framework, we embed motion estimation directly into the convolutional layers to identify and leverage temporal redundancies, enabling significant reductions in redundant computation. However, a key challenge arises due to the structural mismatch between conventional block-based ME and the spatial organization of convolution operations in deep neural networks. Specifically, traditional ME uses large, non-overlapping blocks, whereas convolutional layers operate on overlapping patches (defined by stride) with small, often odd-sized kernels (e.g., $3 \times 3$) for symmetry. 

\begin{figure}[t]
    \centering
    \includegraphics[width=0.95\linewidth]{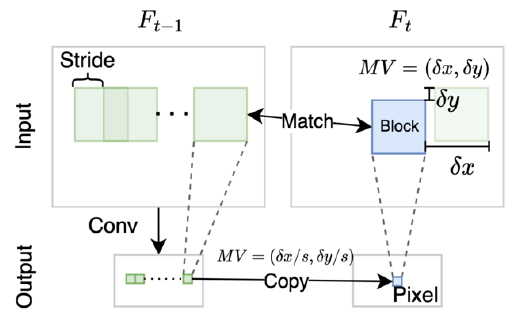}
    \caption{Search pattern of sliding-window motion estimation.}
    \label{fig:search_pattern}
\end{figure}

To bridge this gap, we adopt a \textbf{sliding-window motion estimation strategy}. As shown in Figure~\ref{fig:search_pattern}, the ME block size is matched to the convolution kernel size, and the search is performed in a sliding-window fashion across the input frame. The stride of the convolution is incorporated into the search mechanism, ensuring that each searched block corresponds to a unique output activation. This design aligns the spatial granularity of ME with that of the convolution operation, allowing end-to-end integration.
During ME, the Sum of the absolute difference (SAD) criterion is used to identify the best-matching block from the previous reference frame. The residual between the current block and its match is computed and compared to a threshold $\tau$; any residual value below $\tau$ is suppressed to zero, promoting sparsity. To further reduce unnecessary computation, we introduce an \textbf{early-stopping mechanism}: if the residual sparsity of a candidate match exceeds a predefined threshold (e.g., 0.3), the search is terminated early. SWME follows the same padding and boundary convention as the original convolution layer. During prediction, the MV estimated from the input tensors is rescaled by the layer stride so that warped outputs remain aligned with the corresponding convolutional activations.

\section{Experiments}
We evaluate our framework on three video vision tasks, semantic segmentation, depth estimation, and object detection, and integrate it with convolutional backbones, showing that it is broadly applicable.

\subsection{ISP Power and Latency Measurements}
Table~\ref{tab:isp_power_latency} summarizes published ISP power and latency statistics, showing that ISP processing introduces non-negligible front-end cost. We further measure the system-level latency on our HikRobot prototype, as reported in Table~\ref{tab:hikrobot_latency}. The conventional RGB pipeline incurs 11.08 ms ISP latency and 12.58 ms transmission latency, whereas the proposed ISP-free Bayer pipeline removes ISP processing and reduces transmission latency to 5.41 ms because raw Bayer data is more compact than RGB data.

\begin{table}[H]
\centering
\caption{Summary of published ISP power and latency statistics.}
\label{tab:isp_power_latency}
\small
\setlength{\tabcolsep}{3.5pt}
\resizebox{\linewidth}{!}{%
\begin{tabular}{@{}llll@{}}
\toprule
\textbf{Device / System} & \textbf{Spec} & \textbf{Power [mW]} & \textbf{Latency [ms]} \\
\midrule
\multirow{2}{*}{AP0101CS~\cite{ap0101cs}} &
720p HDR & 129 & -- \\
& 960p HDR & 89 & -- \\
\midrule
\multirow{2}{*}{AP0202AT~\cite{ap0202at}} &
YCbCr@27MHz & 159--202 & -- \\
& \raggedright Spec sheet & 250  & -- \\
\midrule
\multirow{2}{*}{OV492~\cite{ov492}} &
1392$\times$976 & 195 & -- \\
& 1280$\times$720 & 175  & -- \\
\midrule
\multirow{3}{*}{\makecell[l]{VOXL2~\cite{voxl2_latency}\\ (Qualcomm ISP)}} &
1080p$\rightarrow$1080p & -- & 18--20 \\
& 1080p$\rightarrow$720p & -- & 14--16\\
& 4K$\rightarrow$4K & -- & 40--42 \\
\bottomrule
\end{tabular}
}
\end{table}

\begin{table}[H]
\centering
\caption{System-level front end latency measured on our HikRobot prototype.}
\label{tab:hikrobot_latency}
\small
\setlength{\tabcolsep}{4pt}
\resizebox{\linewidth}{!}{%
\begin{tabular}{@{}lcc@{}}
\toprule
\textbf{Pipeline} & \textbf{ISP latency [ms]} & \textbf{Transmission latency [ms]} \\
\midrule
RGB pipeline & 11.08 & 12.58 \\
Bayer pipeline (Ours) & 0 & 5.41 \\
\bottomrule
\end{tabular}
}
\end{table}

\subsection{Semantic Segmentation}
We conduct experiments on two widely used benchmarks: CamVid~\cite{BrostowSFC:ECCV08, BrostowFC:PRL2008} and Cityscapes~\cite{cordts2016cityscapes}. We integrate our framework into two popular convolutional backbones: PSPNet~\cite{zhao2017pyramid} and HRNet~\cite{wang2020deep}, and evaluate their performance in the Bayer domain using our motion-aware formulation. Performance is measured in terms of mean Intersection over Union (mIoU), which reflects segmentation accuracy, and Giga Floating-point Operations (GFLOPs), representing computational cost.

As shown in Table~\ref{tab:result}, our framework achieves segmentation performance comparable to traditional RGB-based models, despite operating directly on raw Bayer data and utilizing motion-aware convolution to reduce computation. For example, on the Cityscapes dataset with HRNet-W48, our method achieves an mIoU of 78.98\%, nearly identical to its RGB counterpart. At the same time, MEVC delivers substantial acceleration, with GFLOPs reduced by approximately 70\% for both PSPNet18 and HRNet-W48.

\begin{table}[htbp]
    \centering
    \caption{Segmentation results with GOP 12, threshold 0.01, and search range 1.}
    \label{tab:result}
    \small
    \setlength{\tabcolsep}{3.5pt}
    \begin{tabular}{@{}llcccc@{}}
        \toprule
        Dataset & Method & \multicolumn{2}{c}{PSPNet18} & \multicolumn{2}{c}{HRNet-W48} \\
        \cmidrule(lr){3-4} \cmidrule(l){5-6}
        & & mIoU(\%) & GFLOPs & mIoU(\%) & GFLOPs \\
        \midrule
        \multirow{2}{*}{CamVid} & RGB  & 69.43 & 309.02 & 75.48 & 247.30 \\
        & Ours & 69.28 & 90.85  & 75.67 & 170.89 \\
        \midrule
        \multirow{2}{*}{Cityscapes} & RGB  & 69.00 & 560.97 & 79.06 & 748.68 \\
        & Ours & 69.10 & 166.52 & 78.98 & 228.91 \\
        \bottomrule
    \end{tabular}
\end{table}

Tables~\ref{tab:performance_camvid} and~\ref{tab:performance_cityscapes} compare our method with existing acceleration approaches on CamVid and Cityscapes, respectively. While previous methods like Accel and AR-Seg apply uniform processing to all frames, our method adapts to temporal redundancy using predictive convolution with motion compensation. BlockCopy and TapLab reduce computation by reusing or propagating blocks, but they do not explicitly correct sparse residuals inside each convolution layer. In contrast, our framework achieves over 70\% reduction in FLOPs with minimal loss in segmentation accuracy.

Notably, the computational savings reported exclude ISP processing cost, which is required for all RGB-domain methods but eliminated in our pipeline, suggesting our reported efficiency gains are conservative.

\begin{table}[htbp]
    \centering
    \caption{Performance comparison with previous works on the CamVid dataset.}
    \label{tab:performance_camvid}
    \begin{tabular}{@{}llll@{}}
        \toprule
        Method & mIoU (\%) & GFLOPs & $\Delta$FLOPs \\
        \midrule
        PSPNet~\cite{zhao2017pyramid} & 69.43 & 309.02 & \textemdash \\
        Accel-DL18~\cite{jain_accel_2019} & 66.15 & 397.70 & +28.7\% \\
        TD\textsuperscript{4}-PSP18~\cite{hu2020temporally} & 70.13 & 363.70 & +17.7\% \\
        BlockCopy~\cite{verelst_blockcopy_2021} & 66.75 & 107.52 & -65.2\% \\
        TapLab-BL2~\cite{feng_taplab_2022} & 67.57 & 117.73 & -61.9\% \\
        Jain et al.~\cite{DBLP:conf/eccv/JainG18} & 67.61 & 146.97 & -52.4\% \\
        AR\textsuperscript{0.6}-PSP18~\cite{hu_efficient_2023} & 70.82 & 101.98 & -67.0\% \\
        \textbf{Ours-PSP18} & 69.28 & 90.85 & -70.6\% \\
        \bottomrule
    \end{tabular}
\end{table}

\begin{table}[htbp]
    \centering
    \caption{Performance comparison with previous works on the Cityscapes dataset.}
    \label{tab:performance_cityscapes}
    \begin{tabular}{@{}llll@{}}
        \toprule
        Method & mIoU (\%) & GFLOPs & $\Delta$FLOPs \\
        \midrule
        PSPNet~\cite{zhao2017pyramid} & 69.00 & 560.97 & \textemdash \\
        Accel-DL18~\cite{jain_accel_2019} & 68.25 & 1011.75 & +80.4\%\\
        TD\textsuperscript{4}-PSP18~\cite{hu2020temporally} & 70.11 & 673.06 & +20.0\% \\
        BlockCopy~\cite{verelst_blockcopy_2021} & 67.69 & 294.20 & -47.6\% \\
        TapLab-BL2~\cite{feng_taplab_2022} & 68.90 & 237.29 & -57.7\% \\
        Jain et al.~\cite{DBLP:conf/eccv/JainG18} & 68.57 & 342.67 & -38.9\% \\
        AR\textsuperscript{0.6}-PSP18~\cite{hu_efficient_2023} & 69.45 & 234.91 & -58.1\% \\
        \textbf{Ours-PSP18} & 69.10 & 166.52 & -70.3\% \\
        \bottomrule
    \end{tabular}
\end{table}

\begin{figure}[hbtp]
    \centering
    \includegraphics[width=0.95\linewidth]{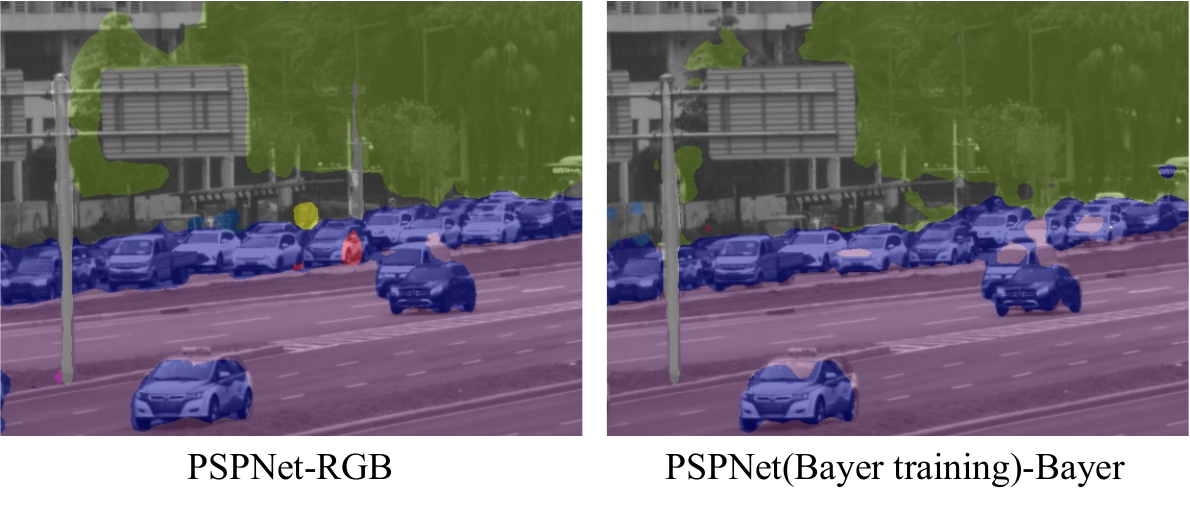}
    \caption{PSPNet on RGB vs. Bayer-PSPNet on Bayer.}
    \label{fig:compare}
\end{figure}

\noindent \textbf{Validation on Real Bayer Data.} Real Bayer data is not human-readable and is difficult to annotate at large scale, restricting quantitative evaluation. However, Figure~\ref{fig:compare} compares RGB PSPNet and Bayer PSPNet on videos captured by our sensors, demonstrating generalization to real Bayer inputs.

\subsection{Depth Estimation}

For monocular depth prediction, we adopt the NYU-Depth V2 dataset~\cite{Silberman:ECCV12} and implement our framework using the fully convolutional architecture of Hu et al.~\cite{hu2019revisiting} as the backbone and adapt it for raw Bayer input using our MEVC-based pipeline.
We evaluate the effectiveness of our proposed MEVC framework on video depth estimation with Root Mean Squared Error (RMSE) and Mean Relative Error (REL). REL is the relative error used by Hu et al.~\cite{hu2019revisiting}, equivalent to mean relative error (MRE). Compared with the RGB baseline, our approach reduces FLOPs by 71.33\% while keeping both RMSE and REL changes below 0.1. This highlights the efficiency of our motion-aware, reconstruction-free framework in preserving depth estimation fidelity under strong acceleration.

\subsection{Object Detection}
To further examine task generality, we deploy MEVC on RetinaNet~\cite{lin2017focal} for object detection on ImageNet-VID~\cite{ILSVRC15}. MEVC achieves 70.48\% FLOPs reduction with only 0.1\% AP degradation, indicating that the motion-compensated convolution design is not limited to dense prediction tasks.

\subsection{Discussions on Parameters}
\noindent \textbf{Group of Pictures (GOP) Length.} We define the Group of Pictures (GOP) in a manner analogous to traditional video coding. Specifically, a GOP consists of a sequence of frames beginning with a reference (key) frame, followed by a set of non-key frames up to (but not including) the next reference frame. In our system, the reference frame is fully processed using conventional convolution, while the subsequent non-key frames are handled using motion-compensated prediction and residual refinement via MEVC.

To examine the effect of GOP size on acceleration and performance, we conduct experiments across a range of GOP lengths. As shown in Table~\ref{tab:GOP}, the overall computational savings generally increase with longer GOPs. This trend arises because more non-key frames are introduced per reference frame, allowing a larger portion of the video to benefit from motion-based acceleration. However, the marginal gain in acceleration diminishes beyond a certain GOP length, due to the increasing temporal distance between non-key frames and their reference, which reduces motion correlation and prediction accuracy.

\begin{table}[htbp]
    \centering
    \caption{Acceleration under different GOP lengths.}
    \label{tab:GOP}
    \begin{tabular}{ccc}
        \toprule
        {GOP length} & {GFLOPs} & {$\Delta$FLOPs (\%)} \\
        \midrule
        2  & 334.45 & -40.4 \\
        6  & 200.13 & -64.3 \\
        10 & 173.07 & -69.2 \\
        12 & 166.52 & -70.3 \\
        \bottomrule
    \end{tabular}
\end{table}

\noindent \textbf{Threshold.} The residual threshold is a critical hyperparameter that governs the sparsity of MEVC’s residual update. Specifically, it determines which residual values are considered negligible and thus omitted from further computation. A higher threshold results in more aggressive filtering of small residuals, reducing the number of pixels processed and thereby improving computational efficiency.

Table~\ref{tab:threshold} presents the relationship between threshold values, FLOPs reduction, and segmentation accuracy (mIoU). As the threshold increases, FLOPs decrease due to fewer residual computations. Notably, when the threshold is set below 0.01, the performance loss remains minimal while computation drops significantly. However, beyond a threshold of 0.05, additional FLOPs savings plateau while the mIoU begins to degrade more noticeably. This suggests that the majority of residual noise lies below the 0.01 level, and overly aggressive thresholding risks discarding informative signal in the residual map.

\begin{table}[htbp]
    \centering
    \caption{Acceleration under different thresholds.}
    \label{tab:threshold}
    \begin{tabular}{cccc}
        \toprule
        {Threshold} & {mIoU (\%)} & {GFLOPs} & {$\Delta$FLOPs (\%)} \\
        \midrule
        0    &  69.10 & 325.09 & -41.9 \\
        0.01 &  69.10 & 166.52 & -70.3 \\
        0.05 &  66.06 &  74.11 & -86.8 \\
        0.1  &  65.74 &  71.67 & -87.2 \\
        \bottomrule
    \end{tabular}
\end{table}

\subsection{Ablation Study}
To demonstrate the effectiveness of each component in our framework, Table~\ref{tab:ablation} reports acceleration and segmentation performance under four settings: 1) PSPNet processes all frames in Bayer videos; 2) MEVC without compensation; 3) MEVC with compensation but without thresholding; and 4) MEVC with both compensation and thresholding.

\begin{table}[H]
    \centering
    \caption{Acceleration and performance under different settings.}
    \label{tab:ablation}
    \begin{tabular}{cccc}
        \toprule
        {Setting} & {mIoU (\%)} & {GFLOPs} & {$\Delta$FLOPs (\%)} \\
        \midrule
        1 & 69.00 & 560.97 & -0.0 \\
        2 & 41.74 &  72.41 & -87.1  \\
        3 & 69.10 & 325.09 & -41.9  \\
        4 & 69.10 & 166.52 & -70.3  \\
        \bottomrule
    \end{tabular}
\end{table}

\section{Conclusion}
We present a unified framework that rethinks the design of video computer vision systems by eliminating the ISP and integrating motion estimation directly into the convolutional process. By operating on raw Bayer data and introducing Motion Estimation-based Video Convolution (MEVC), our method effectively reduces redundant computation through motion-compensated prediction and sparse residual refinement. A sliding-window alignment strategy bridges the gap between block-based motion estimation and convolutional feature extraction, enabling seamless integration into standard CNN backbones. Experiments across semantic segmentation, depth estimation, and object detection demonstrate that our approach achieves over 70\% FLOPs reduction with minimal performance loss. This framework is broadly applicable to convolution-based neural networks and downstream tasks. 

\small
\bibliographystyle{IEEEbib}
\bibliography{strings,refs}

\end{document}